\documentclass[10pt,twocolumn,letterpaper]{article}

\usepackage{wacv}
\usepackage{times}
\usepackage{epsfig}
\usepackage{graphicx}
\usepackage{amsmath}
\usepackage{amssymb}
\usepackage{makecell}
\usepackage{multirow}
\usepackage[dvipsnames]{xcolor} 
\usepackage[linesnumbered,ruled,vlined]{algorithm2e}
\usepackage{arydshln}
\usepackage{hyperref}

\SetCommentSty{mycommfont} 

\wacvfinalcopy 

\def\wacvPaperID{****} 

\ifwacvfinal\fi
\begin{document}

\title{Few-Shot Scene Adaptive Crowd Counting Using Meta-Learning}


\author{Mahesh Kumar Krishna Reddy \hspace{0.5cm} Mohammed Asiful Hossain \hspace{0.5cm} Mrigank Rochan \hspace{0.5cm}  Yang Wang \\
University of Manitoba \hspace{0.5cm}\\
{\tt\small \{kumarkm, hossaima, mrochan, ywang\}@cs.umanitoba.ca} \hspace{1cm}
}


\maketitle
\ifwacvfinal\thispagestyle{empty}\fi

\begin{abstract}
   We consider the problem of {\normalfont{few-shot scene adaptive crowd counting}}. Given a target camera scene, our goal is to adapt a model to this specific scene with only a few labeled images of that scene. The solution to this problem has potential applications in numerous real-world scenarios, where we ideally like to deploy a crowd counting model specially adapted to a target camera. We accomplish this challenge by taking inspiration from the recently introduced learning-to-learn paradigm in the context of few-shot regime. In training, our method learns the model parameters in a way that facilitates the fast adaptation to the target scene. At test time, given a target scene with a small number of labeled data, our method quickly adapts to that scene with a few gradient updates to the learned parameters. Our extensive experimental results show that the proposed approach outperforms other alternatives in few-shot scene adaptive crowd counting. Code is available at \url{https://github.com/maheshkkumar/fscc}
   
\end{abstract}


\section{Introduction}\label{sec:introduction}

Recently, the problem of crowd counting~\cite{li2018csrnetcounting, sam2017switchingcounting,sindagi2017generatingcounting, zhang2015crosscounting, zhang2016singlecounting} is drawing increasing attention in computer vision research. The key reason for this surge in interest is the demand of automated complex crowd scene understanding that appears in computer vision applications such as surveillance, traffic monitoring, etc. Although the contemporary methods for crowd counting are promising, they have some significant limitations. One main limitation of existing methods is that it is hard to adapt them to a new crowd scene. This is due to the fact that these methods typically require a large number of labeled training data which is expensive and time-consuming to obtain. In this paper, we focus on this issue and propose a method that learns to adapt to a new crowd scene with very few labeled examples of that scene. 




\begin{figure}[!t]
	\centering
	\includegraphics[width=11.5cm,height=5.0cm]{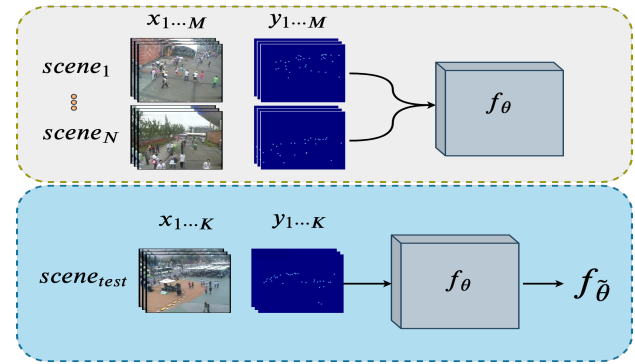}
	\caption{Illustration of our problem setting. (Top row) During training, we have access to a set of $N$ different camera scenes where each scene comes with $M$ labeled examples. From such training data, we learn the model parameters $\theta$ of a mapping function $f_{\theta}$ such that $\theta$ is generalizable across scenes in estimating the crowd count. (Bottom row) Given a test (or target) scene, we assume that we have a small number of $K$ labeled images from this scene, where $K \ll M$ (e.g., $K \in \{1, 5\}$) to learn the scene-specific parameters $\tilde{\theta}$. With the help of meta-learning guided approach we quickly adapt $f_{\theta}$ to $f_{\tilde{\theta}}$ that predicts more accurate crowd count than other alternative solutions.}
	\label{fig:intro_diagram}
\end{figure}

Most current approaches~\cite{li2018csrnetcounting,sam2017switchingcounting,sindagi2017generatingcounting, zhang2015crosscounting, zhang2016singlecounting} of crowd counting treat it as a supervised regression problem where a model is learned to produce a crowd density map for the given image. In the training phase, the model learns to predict the density map of an input image given its ground-truth crowd density map as the label. The final crowd count is obtained by summing over the pixels in the estimated density map. Once the model is learned, it can be used to estimate the crowd count in test images. The main drawback of existing approaches is that they produce a single learned model that will be used in all unseen images. In order to make the model generalize well, we often need to make sure that the labeled training data is diverse enough to cover all possible scenarios which is infeasible. 


A recent work~\cite{hossain2019one} argues that it is more effective to learn and deploy a model specifically tuned to a particular scene, instead of learning a generic model that hopefully works well in all scenes. Let us consider the video surveillance scenario. Once a surveillance camera is installed, the images captured by the camera are constrained mainly by the camera parameters and the 3D geometry of a specific scene. From the viewpoint of practical applications, we do not need the crowd counting model to perform well on arbitrary images. Instead, we only need the model to be tuned to this particular scene. Of course, if we can get access to adequate labeled training images from this camera, a simple solution is to train a model for this scene using its training images. However, this is unrealistic since it requires collecting a large number of labeled images from the target scene whenever a new surveillance camera is installed. Moreover, generating adequate labeled data for a specific camera scene can be expensive and tedious. Ideally, we would like a way of adapting a model to work well in a new camera scene with only a few labeled examples from that scene. 


We consider the few-shot scene adaptive crowd counting similar to \cite{hossain2019one}. During training, we have access to a set of training images from different scenes (e.g., each scene might correspond to one specific camera installed at one particular location). During testing, we have a new target crowd scene to which we want to adapt our model. Moreover, we consider that we have a small number (e.g., $1$ or $5$) of labeled images from this target scene. During training, we learn optimal (generalizable) model parameters from multiple scene-specific data by considering few-labeled images per scene. During testing, we consider the learned parameters to be a good initial point to adapt to a specific new scene. To be precise, we aim at learning the generalizable model parameters in a fashion that it produces more accurate performance when adapting to a new target scene with few gradient descent steps provided only a few labeled images from the target scene. Figure~\ref{fig:intro_diagram} shows an illustration of the problem in this paper. We address the proposed few-shot crowd counting problem using meta-learning~\cite{finn2017modelagnostic} that is capable of fast adaptation to new camera scenes. 

This paper makes the following contributions. First, we propose a meta-learning inspired approach to solve the few-shot scene adaptive crowd counting problem. Using the meta-learning, the model parameters are learned in a way that facilitates effective fine-tuning to a new scene with a few labeled images. Previous work in \cite{hossain2019one} uses a fine-tuning approach for this problem. The limitation of this fine-tuning approach is that it can only update certain layers that are closer to the output in the decoder to a target scene. In contrast, our approach does not have such limitation and can be used to adapt any parameters in the decoder. Second, we perform a thorough evaluation of the performance of our proposed approach on several benchmark datasets and show that the method outperforms other alternative baselines. Our approach also outperforms the fine-tuning approach in \cite{hossain2019one}.

\section{Related Work}\label{sec:related_work}
\noindent {\bf Crowd Counting}: The research in crowd counting can be grouped into either detection~\cite{dalal2005histogramscounting, dollar2012pedestriandetection}, regression~\cite{chan2009bayesiancounting, idrees2013multicounting} or density-based~\cite{lempitsky2010learningcounting, pham2015counting} methods as proposed by~\cite{loy2013crowdcounting}. Earlier work focuses on the detection and regression-based approaches. In recent years, density-based approaches using deep learning models have become popular and show superior performance. Zhang \textit{et al}.~\cite{zhang2015crosscounting} propose an approach with two learning objectives for density estimation and crowd counting. Additionally, they propose a non-parametric method to fine-tune the model to minimize the distribution difference between the source and target scenes. Zhang \textit{et al}.~\cite{zhang2016singlecounting} address crowd counting by proposing a multi-column neural network to handle an input image at multiple scales to overcome the problem of scale variations. Sam et \textit{al}.~\cite{sam2017switchingcounting} propose to estimate the density of an image patch from a regressor selected based on the density level classifier. Sindagi and Patel~\cite{sindagi2017generatingcounting} propose to encode both local and global input image contexts to estimate the density map. In this paper, our backbone crowd counting architecture is based on~\cite{li2018csrnetcounting}, since it has been shown to achieve state-of-the-art performance.

In the context of crowd counting adaptation, Loy \textit{et al}.~\cite{change2013semicounting} propose a non-CNN semi-supervised adaptation method by exploiting unlabeled data in the target domain. The drawback of this approach is that it requires corresponding samples that have common labels between the source and target domains. This information is usually not available in recent crowd counting datasets. Wang \textit{et a}l.~\cite{wang2019learning} propose to generate a large synthetic dataset and perform domain adaptation to the real-world target domain. One drawback of this method is that it requires the prior knowledge about the distribution of the target domain in order to manually select the scenes in the synthetic dataset. Hossain \textit{et al.}~\cite{hossain2019one} propose a one-shot adaptation approach based on fine-tuning few layers in the decoder network for adapting a crowd counting model to a specific scene.

\begin{figure*}[!t]
	\centering
	\includegraphics[width=\textwidth]{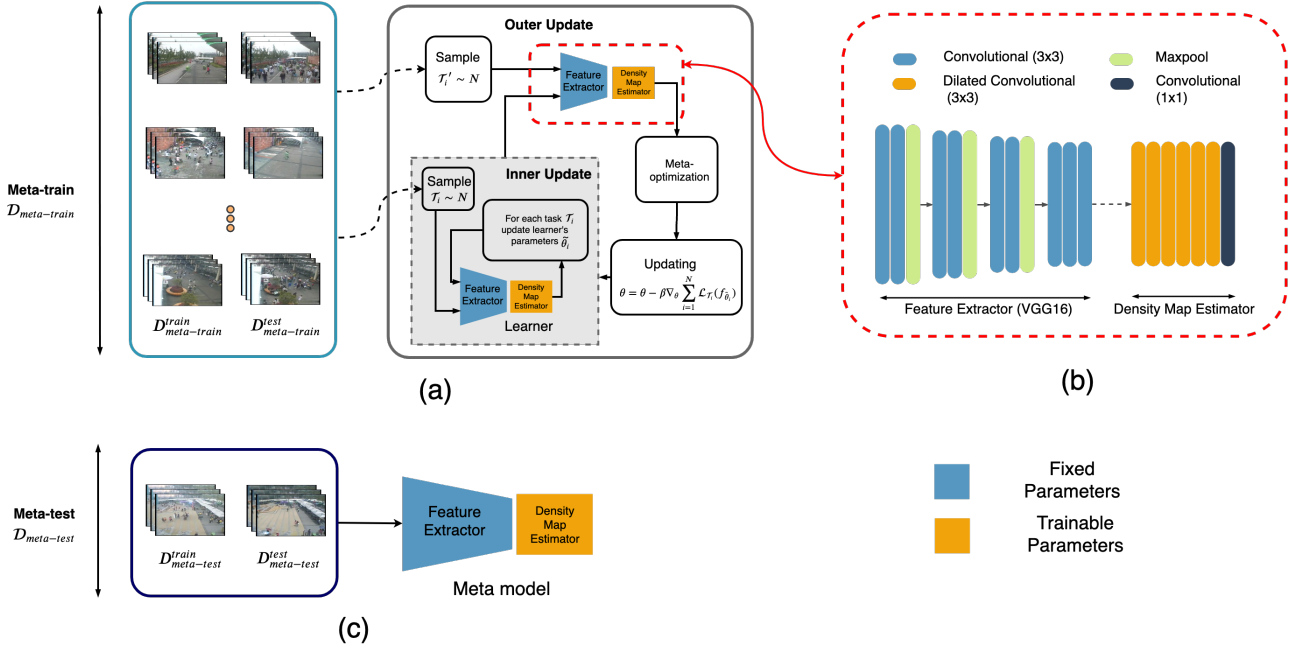}
	\caption{An overview of the main components of our model. (a) \textit{Meta-training} stage on $\mathcal{D}_{meta-train}$. The meta-training involves optimizing an inner-update over each scene and an outer-update across different scenes. (b) Backbone crowd counting network. We use the CSRNet~\cite{li2018csrnetcounting} as the backbone architecture. It comprises of a feature extractor and a density map estimator. (c) \textit{Meta-testing} on $\mathcal{D}_{meta-test}$. We adapt the trained meta-model with $\theta$ to a new target scene by fine-tuning on $K$ images from this scene and test on other images from this scene.}
	\label{fig:learning}
\end{figure*}

\noindent {\bf Few-Shot Learning}: The goal of few-shot learning is to learn a model from limited training examples for a task. Previously, Li \textit{et al}.~\cite{feifei06_pami_oneshot} propose a method for unsupervised one-shot learning by casting the problem in a probabilistic setting. Lake \textit{et al}.~\cite{lake2013one} use compositionality and causality for one-shot scenario through Hierarchical Bayesian learning system. Luo \textit{et al}.~\cite{luo2017label} demonstrate the transferability of representations across domains with few labeled data. A different perspective to tackle few-shot learning is by treating it is as a meta-learning problem (also known as \textit{learning to learn}~\cite{bengio1992optimization, schmidhuber1987evolutionary}). The essence of using meta-learning for few-shot learning problem involves a neural network as a learner to learn about a new task with just a few instances. The recent work in meta-learning can be grouped into metric-based~\cite{koch2015siamese, snell2017prototypicalnetworks, sung18_cvpr, vinyals2016matchingnetworks}, model-based~\cite{munkhdalai2017meta, santoro2016metanetworks} or optimization-based~\cite{finn2017modelagnostic, nichol2018first, ravi2016optimizationmethod}. The metric-based~\cite{koch2015siamese, snell2017prototypicalnetworks, sung18_cvpr, vinyals2016matchingnetworks} methods in general learn a distance function to measure the similarity between data points belonging to the same class. Memory or model-based~\cite{munkhdalai2017meta, santoro2016metanetworks} approaches employ a memory component to store previously used training examples. The optimization-based~\cite{finn2017modelagnostic, nichol2018first, ravi2016optimizationmethod} frameworks learn good initialization parameters based on learning from multiple tasks that favour fast adaptation on a new task. The above works primarily target image recognition challenge, in our proposed work we follow the optimization-based meta-learning mechanism similar to~\cite{finn2017modelagnostic} for a more challenging problem of crowd density estimation as it has shown to achieve superior performance compared to other optimization based methods. 


\section{Few-shot Scene Adaptive Crowd Counting}\label{sec:approach}

In this section, we first describe the problem setup for few-shot scene adaptive crowd counting~(Sec.~\ref{sec:formulation}). We then introduce our proposed approach for scene adaptive crowd counting using meta-learning~(Sec.~\ref{sec:meta_learning}). 



\noindent \subsection{Problem Setup}\label{sec:formulation}
We describe how we formulate the scene adaptive crowd counting as a few-shot learning problem using meta-learning. In a traditional supervised machine learning setting, we are given a dataset $\mathcal{D}=\{D^{train}, D^{test}\}$, where $D^{train}$ and $D^{test}$ are the training and test sets, respectively. The goal is to learn a mapping function $f_{\theta}: x \rightarrow y$ that maps an input $x$ (e.g. an input image) to its corresponding label $y$ (e.g. the crowd density map). We use $\theta$ to denote the parameters of the mapping function $f_{\theta}$. We learn $\theta$ by optimizing its corresponding loss function defined on $D^{train}$. After training, we test the generalization of the learned model $f_{\theta}$ on $D^{test}$.

In contrast, a few-shot meta-learning model is trained on a set of $N$ tasks during meta-learning (\textit{meta-training}) from $\mathcal{D}_{meta-train}$, where each task has its training and test sets. We use $\mathcal{T}_i=\{D_i^{train}, D_i^{test}\}$ ($i=1,2,...,N$), where $\mathcal{T}_{i} \in \mathcal{D}_{meta-train}$ to denote the $i$-th task (also called episode) during the meta-learning phase. The notations $D_i^{train}$ and $D_i^{test}$ correspond to the training set and the test set of the $i$-th task, respectively. Note that during the meta-learning phase, both $D_i^{train}$ and $D_i^{test}$ consist of labeled examples. We consider each camera scene as a task in the meta-learning formulation. Each of the training $i$-th scene consists of $M$ labeled images. However, in our work, we randomly sample a small number $K \in \{1, 5\}$ and $K \ll M$ labeled images for the $i$-th scene in each learning iteration to form $D_i^{train}$. The $D_i^{test}$ is the test set for the $i$-th scene. This setup reflects the real-world problem of having to learn from a few labeled images. Our goal of the meta-learning is to learn the model in a way that it can adapt to a new scene using only a few training examples from the new scene. During testing (i.e., \textit{meta-testing}) on $\mathcal{D}_{meta-test}$, we are given a new target scene $\mathcal{T}_{new}=\{D_{new}^{train}, D_{new}^{test}\}$, where $D_{new}^{train}$ consists of a few (e.g. $K$) labeled images from the target scene. The goal is to quickly adapt the model using $D_{new}^{train}$ so that the adapted model performs well on $D_{new}^{test}$ which is the test data for this target scene. In our work, we use the meta-learning approach in \cite{finn2017modelagnostic} called \emph{MAML}. MAML learns a set of initial model parameters during the meta-training stage. The model parameters learned during \textit{meta-training} are used for initializing the model during \textit{meta-testing} and is later fine-tuned on the few examples from a new target task. The adapted model with fine-tuned parameters is expected to perform well on the test images from the target task.

\noindent \subsection{Our Approach} \label{sec:meta_learning}

Consider a crowd counting model $f_{\theta}$ with the model parameters $\theta$. Given an input image $x$, the output of $f_{\theta}(x)$ is a crowd density map representing the density level at different spatial locations in the image. The crowd count can be obtained by summing over entries in the generated density map. When learning to adapt to a particular scene $\mathcal{T}_i$, the model parameters are updated using a few gradient steps to optimize the loss function defined on $D_{i}^{train}$. This learning step can be considered as \textit{inner-update} during meta-learning and the optimization is expressed as follows:  
\begin{equation}
\begin{aligned}
	\tilde{\theta_{i}} &= \theta - \alpha \nabla_{\theta} \mathcal{L}_{\mathcal{T}_{i}}(f_{\theta}) \\
	\textrm{where }
	\mathcal{L}_{\mathcal{T}_{i}}(f_{\theta}) &= \sum_{(x^{(j)}, y^{(j)}) \in D_{i}^{train}} \|f_{\theta}(x^{(j)}) - y^{(j)}\|_{F}^{2} 
	\label{eq:inner_update}
\end{aligned}
\end{equation}
\noindent where $x^{(j)}$ and $y^{(j)}$ denote a training image and its corresponding ground-truth density map from the scene $\mathcal{T}_i$, respectively. We use $||\cdot||$ to denote the Frobenius norm that measures the difference between the predicted crowd density map $f_{\theta}(x^{(j)})$ and the ground-truth density map $y^{(j)}$. Here $\alpha$ is the learning rate in the \textit{inner-update} and its value is fixed in our implementation. We then define a loss function on $D_{i}^{test}$ using $\tilde{\theta_{i}}$ as follows:

\begin{equation}
  \mathcal{L}_{\mathcal{T}_i}(f_{\tilde{\theta_{i}}})=\sum_{(x^{(j)}, y^{(j)}) \in D_{i}^{test}}\|f_{\tilde{\theta_{i}}}(x^{(j)}) - y^{(j)}\|_{F}^{2} 
  \label{eq:meta_optimization}
\end{equation}


During the meta-learning phase, we learn the model parameters $\theta$ by optimizing $\mathcal{L}_{\mathcal{T}_i}(f_{\tilde{\theta_{i}}})$ across $N$ different training scenes. This will effectively learn $\theta$ in a way that when we update $\theta$ with a few gradient steps in a new scene, the updated parameters $\tilde{\theta}$ will perform well on test images from this scene. This optimization problem (or \textit{outer-update}) is similar to the optimization described in~\cite{finn2017modelagnostic} and it is expressed as:

\begin{equation}
	\theta = \theta - \beta \nabla_{\theta} \sum_{i=1}^N \mathcal{L}_{\mathcal{T}_i}(f_{\tilde{\theta_{i}}})
	\label{eq:outer_update}
\end{equation}

Fig.~\ref{fig:learning} shows an illustration of this meta-learning inspired process. The result of the meta-learning phase is the set of model parameters $\theta$. Given a new scene, we use $\theta$ to initialize the model and obtain the scene adaptive parameters $\tilde{\theta}$ by fine-tuning the parameters on the few examples from the target scene with a few gradient updates. The intuition is that a well-learned parameters $\theta$ should be able to generalize to new scenes with only a few gradient updates. In our implementation for few-shot scene adaptive crowd counting, we compute the second derivatives to optimize Eq.~\ref{eq:outer_update} during outer-update as described in~\cite{finn2017modelagnostic}. \\
 


\noindent {\bf Backbone Network Architecture}: Our proposed few-shot learning approach for crowd density estimation can be used with any backbone crowd counting network architecture. In this paper, we use the CSRNet~\cite{li2018csrnetcounting} (see Fig.~\ref{fig:learning}) as our backbone network since it has shown to achieve state-of-the-art performance in crowd counting. The network consists of a feature extractor and a density map estimator. The feature extractor uses VGG-16~\cite{Simonyan14c} to extract a feature map of the input image. Following \cite{li2018csrnetcounting}, we use the first 10 layers (up to Conv4$\_3\_3$) of VGG-16 as the feature extractor. The output of the feature extractor has a resolution of $1/8$ of the input image. The density map estimator consists of a series of dilated convolutional layers~\cite{yu2016multi} to regress the output crowd density map for the given image.


We use a pre-trained VGG-16~\cite{Simonyan14c} model on ImageNet~\cite{Deng2009ImageNetAL} to initialize the weights of the feature extractor part of our network. The weights of the dilated convolutional layers in the density map estimator part of the network are initialized from a Gaussian with 0.01 standard deviation. We then train the network end-to-end on the training set of WorldExpo'10~\cite{zhang2015crosscounting} dataset to learn how to produce a density map for an image containing the crowd. We refer to this trained network as ``\textit{Baseline pre-trained}'' in the remaining of the paper. Note that although the baseline pre-trained model is learned on data from multiple training scenes, it is susceptible when used for adaptation in few labeled data regime as it is not specifically designed to learn from few images which we discuss in the later section. Therefore, in order to overcome this limitation, we use this baseline pre-trained network as the initialization for the meta-learning phase. During meta-learning, we fix the parameters of the feature extractor and train only density map estimator on different scene-specific data. We follow the training scheme described in this section to learn to adapt to a scene with a few labeled images.

\begin{table*}[!ht]
  \centering
  \footnotesize
  \resizebox{0.9\textwidth}{!}{%
\def\arraystretch{1.1}
\begin{tabular}{cccccccc}
\Xhline{2\arrayrulewidth}
\multirow{2}{*}{\textbf{Target}} & \multirow{2}{*}{\textbf{Methods}} & \multicolumn{3}{c}{\textbf{1-shot (K=1)}}                                         & \multicolumn{3}{c}{\textbf{5-shot (K=5)}}                                         \\ 
                                 &                                   & \multicolumn{1}{c}{MAE}                       & \multicolumn{1}{c}{RMSE}                       & \multicolumn{1}{c}{MDE}                        & \multicolumn{1}{c}{MAE}                       & \multicolumn{1}{c}{RMSE}                       & \multicolumn{1}{c}{MDE}                        \\ \cline{1-8} 
\multirow{5}{*}{Scene 1}         
& Baseline pre-trained                       & 5.55                      & 6.31                      & 0.70                       & 5.55                      & 6.31                      & 0.70                       \\ 
& Baseline fine-tuned                        & 5.45 $\pm$ {\footnotesize 0.03}           & 6.23 $\pm$ {\footnotesize 0.03}           & 0.68 $\pm$ {\footnotesize 0.004}           & 5.06 $\pm$ {\footnotesize 0.11}           & 5.88 $\pm$ {\footnotesize 0.10}           & 0.63 $\pm$ {\footnotesize 0.005}           \\
& Meta pre-trained             & 4.63           & 5.5 & 0.529 &  4.63           & 5.5 & 0.529 \\
                                 & \textbf{Ours w/o ROI}             & 3.47 $\pm$ {\footnotesize 0.01}           & \textbf{4.19 $\pm$ {\footnotesize 0.01}}  & 0.50 $\pm$ {\footnotesize 0.007}           & 3.42 $\pm$ {\footnotesize 0.03}           & 4.81 $\pm$ {\footnotesize 0.007}          & \textbf{0.29 $\pm$ {\footnotesize 0.004}}   \\ 
                                 & \textbf{Ours w/ ROI}            & \textbf{3.19 $\pm$ {\footnotesize 0.03}}  & 4.30 $\pm$ {\footnotesize 0.07}           & \textbf{0.38 $\pm$ {\footnotesize 0.03}}   & \textbf{3.05 $\pm$ {\footnotesize 0.06}}  & \textbf{4.19 $\pm$ {\footnotesize 0.15}}  & 0.31 $\pm$ {\footnotesize 0.08}            \\ \hline
\multirow{5}{*}{Scene 2}          
& Baseline pre-trained                       & 24.07                     & 34.29                     & 0.17                       & 24.07                     & 34.29                     & 0.17                       \\ 
& Baseline fine-tuned                        & 22.74 $\pm$ {\footnotesize 0.47}          & 32.92 $\pm$ {\footnotesize 0.66}          & 0.15 $\pm$ {\footnotesize 0.003}           & 20.84 $\pm$ {\footnotesize 1.03}          & 30.49 $\pm$ {\footnotesize 1.37}          & 0.156 $\pm$ {\footnotesize 0.001}          \\
& Meta pre-trained             & 21.65 & 30.51 & 0.185 & 21.65 & 30.51 & 0.185 \\
& \textbf{Ours w/o ROI}             & 12.05 $\pm$ {\footnotesize 0.74}          & 16.62 $\pm$ {\footnotesize 1.10}          & 0.11 $\pm$ {\footnotesize 0.007}           & 11.41 $\pm$ {\footnotesize 0.54}          & 15.35 $\pm$ {\footnotesize 0.51}          & 0.11 $\pm$ {\footnotesize 0.015}           \\
& \textbf{Ours w/ ROI}            & \textbf{11.17 $\pm$ {\footnotesize 1.01}} & \textbf{15.50 $\pm$ {\footnotesize 1.18}} & \textbf{0.11 $\pm$ {\footnotesize 0.012}}  & \textbf{10.73 $\pm$ {\footnotesize 0.36}} & \textbf{14.95 $\pm$ {\footnotesize 0.60}} & \textbf{0.10 $\pm$ {\footnotesize 0.003}}  \\ \hline
\multirow{5}{*}{Scene 3}         
& Baseline pre-trained                       & 35.54                     & 40.78                     & 0.40                       & 35.54                     & 40.78                     & 0.40                       \\
& Baseline fine-tuned                        & 33.89 $\pm$ {\footnotesize 0.26}          & 39.33 $\pm$ {\footnotesize 0.25}          & 0.38 $\pm$ {\footnotesize 0.03}            & 31.05 $\pm$ {\footnotesize 0.41}          & 36.70 $\pm$ {\footnotesize 0.43}          & 0.34 $\pm$ {\footnotesize 0.004}           \\
& Meta pre-trained   & 36.18 & 42.32 & 0.402 & 36.18 & 42.32 & 0.402 \\
& \textbf{Ours w/o ROI}             & 8.15 $\pm$ {\footnotesize 0.17}           & 11.04 $\pm$ {\footnotesize 0.42}          & \textbf{0.09 $\pm$ {\footnotesize 0.04}}  & 8.31 $\pm$ {\footnotesize 0.54}           & \textbf{10.75 $\pm$ {\footnotesize 0.54}} & 0.10 $\pm$ {\footnotesize 0.009}          \\
& \textbf{Ours w/ ROI}            & \textbf{8.07 $\pm$ {\footnotesize 0.23}}  & \textbf{10.92 $\pm$ {\footnotesize 0.21}} & 0.10 $\pm$ {\footnotesize 0.007}           & \textbf{8.18 $\pm$ {\footnotesize 0.24}}  & 10.96 $\pm$ {\footnotesize 0.31}          & \textbf{0.09 $\pm$ {\footnotesize 0.002}} \\ \hline
\multirow{5}{*}{Scene 4}         
& Baseline pre-trained                       & 23.95                     & 28.57                     & 0.19                       & 23.95                     & 28.57                     & 0.19                       \\ 
& Baseline fine-tuned                        & 15.69 $\pm$ {\footnotesize 0.28}          & 18.96 $\pm$ {\footnotesize 0.27}          & 0.14 $\pm$ {\footnotesize 0.003}           & 16.67 $\pm$ {\footnotesize 0.10}          & 19.70 $\pm$ {\footnotesize 0.16}          & 0.15 $\pm$ {\footnotesize 0.002}           \\
& Meta pre-trained             & 22.44 & 28.25 & 0.183 & 22.44 & 28.25 & 0.183 \\
& \textbf{Ours w/o ROI}             & 9.74 $\pm$ {\footnotesize 0.09}          & 11.9 $\pm$ {\footnotesize 0.12}          & 0.084 $\pm$ {\footnotesize 0.001}           & 11.21 $\pm$ {\footnotesize 0.47}          & 16.1 $\pm$ {\footnotesize 0.45}          & 0.118 $\pm$ {\footnotesize 0.004}           \\ 
& \textbf{Ours w/ ROI}            & \textbf{9.39 $\pm$ {\footnotesize 0.26}} & \textbf{11.78 $\pm$ {\footnotesize 0.34}} & \textbf{0.07 $\pm$ {\footnotesize 0.02}}  & \textbf{9.41 $\pm$ {\footnotesize 0.21}} & \textbf{11.91 $\pm$ {\footnotesize 0.17}} & \textbf{0.08 $\pm$ {\footnotesize 0.002}}   \\ \hline
\multirow{5}{*}{Scene 5}         
& Baseline pre-trained                       & 10.70                     & 13.0                      & 0.67                       & 10.70                     & 13.0                      & 0.67                       \\  
 & Baseline fine-tuned                        & 8.9 $\pm$ {\footnotesize 0.05}            & 11.7 $\pm$ {\footnotesize 0.04}           & 0.50 $\pm$ {\footnotesize 0.03}            & 7.79 $\pm$ {\footnotesize 0.35}           & 10.57 $\pm$ {\footnotesize 0.66}          & 0.44 $\pm$ {\footnotesize 0.015}           \\
& Meta pre-trained            & 9.78 & 12.26 & 0.605 & 9.78 & 12.26 & 0.605\\
& \textbf{Ours w/o ROI}             & 4.09 $\pm$ {\footnotesize 0.01}           & 7.36 $\pm$ {\footnotesize 0.01}           & 0.196 $\pm$ {\footnotesize 0.001}           & 4.28 $\pm$ {\footnotesize 0.14}           & 7.68 $\pm$ {\footnotesize 0.60}           & 0.20 $\pm$ {\footnotesize 0.001}           \\  
& \textbf{Ours w/ ROI}            & \textbf{3.82 $\pm$ {\footnotesize 0.05}}  & \textbf{6.91 $\pm$ {\footnotesize 0.11}}  & \textbf{0.192 $\pm$ {\footnotesize 0.001}} & \textbf{3.91 $\pm$ {\footnotesize 0.26}}  & \textbf{7.18 $\pm$ {\footnotesize 0.85}}  & \textbf{0.18 $\pm$ {\footnotesize 0.001}} \\ \hline
\multirow{5}{*}{\textbf{Average}}                                    
& Baseline pre-trained                       & 19.96                     & 24.59                      & 0.42                       & 19.96                     & 24.59                      & 0.42                       \\  
& Baseline fine-tuned                        & 17.33 & 21.82            & 0.37 & 16.28           & 20.66            & 0.34                         \\
& Meta pre-trained             & 18.93 & 23.76 & 0.38 & 18.93 & 23.76 & 0.38 \\ 
& \textbf{Ours w/o ROI}             & 7.5                 & 10.22                 & 0.197                  & 7.7                 & 10.93                 & 0.165                  \\ 
& \textbf{Ours w/ ROI}            & \textbf{7.12}                      & \textbf{9.88}                         & \textbf{0.172 }                          & \textbf{7.05                         } & \textbf{9.83}                           & \textbf{0.155} \\ \Xhline{2\arrayrulewidth}
\end{tabular}}
\caption{Results on WorldExpo'10~\cite{zhang2015crosscounting} test set with $K = 1$ and $K = 5$ train images in the targe scene. We report the performance our our approach with and without ROI. We also compare with three baselines \textit{Baseline pre-trained}, \textit{Baseline fine-tuned} and \textit{Meta pre-trained}. We compare the results across 5 test scenes and the last two rows represent the average score for our models.}
\label{tab:worldexpo}
\end{table*}

\section{Experiments}\label{sec:experiments}
In this section, we first introduce the datasets and experiment setup~(Sec.~\ref{sec:datasets}). We then describe several baselines for comparison~(Sec.~\ref{sec:baselines}). We present the experimental results~(Sec.~\ref{sec:experimental_results}).

\begin{table*}[!ht]
\centering
\footnotesize
\def\arraystretch{1.1}
\begin{tabular}{ccccccc}
\Xhline{2\arrayrulewidth}
\multirow{2}{*}{\textbf{Methods}} & \multicolumn{3}{c}{\textbf{1-shot (K=1)}}            & \multicolumn{3}{c}{\textbf{5-shot (K=5)}}            \\
                                  & \multicolumn{1}{c}{MAE}             & \multicolumn{1}{c}{RMSE}             & \multicolumn{1}{c}{MDE}               & \multicolumn{1}{c}{MAE}             & \multicolumn{1}{c}{RMSE}             & \multicolumn{1}{c}{MDE}               \\ \cline{1-7}
Baseline pre-trained                       & 7.29            & 7.96            & 0.22              & 7.29            & 7.96            & 0.22              \\ 
Baseline fine-tuned                        & 7.11 $\pm$ {\footnotesize 0.09} & 7.80 $\pm$ {\footnotesize 0.08} & 0.21 $\pm$ {\footnotesize 0.003}  & 6.58 $\pm$ {\footnotesize 0.07} & 7.32 $\pm$ {\footnotesize 0.06} & 0.20 $\pm$ {\footnotesize 0.002}  \\ 
 Meta pre-trained & 7.01 & 7.69 & 0.230 & 7.01 & 7.69 & 0.230 \\
\textbf{Ours w/o ROI}                    & 2.52 $\pm$ 0.08 & 3.26 $\pm$ {\footnotesize 0.12} & 0.078 $\pm$ {\footnotesize 0.002} & 2.53 $\pm$ {\footnotesize 0.18} & 3.25 $\pm$ {\footnotesize 0.27} & 0.078 $\pm$ {\footnotesize 0.004} \\ 
\textbf{Ours w/ ROI}                   & \textbf{2.44 $\pm$ {\footnotesize 0.02}} & \textbf{3.12 $\pm$ {\footnotesize 0.03}} & \textbf{0.076 $\pm$ {\footnotesize 0.001}}  & \textbf{2.37 $\pm$ {\footnotesize 0.02}} & \textbf{3.04 $\pm$ {\footnotesize 0.01}} & \textbf{0.073 $\pm$ {\footnotesize 0.001}}  \\ \Xhline{2\arrayrulewidth}
\end{tabular}
\caption{Results on the Mall~\cite{change2013semicounting} dataset with $K=1$ and $K=5$ images in the target scene. The meta-training is performed on the WorldExpo'10 training data.}
\label{tab:worldexpo_mall}
\end{table*}

\begin{table*}[!ht]
\centering
\footnotesize
\def\arraystretch{1.1}
\begin{tabular}{ccccccc}
\Xhline{2\arrayrulewidth}
\multirow{2}{*}{\textbf{Methods}} & \multicolumn{3}{c}{\textbf{1-shot (K=1)}}            & \multicolumn{3}{c}{\textbf{5-shot (K=5)}}            \\
                                  & \multicolumn{1}{c}{MAE}             & \multicolumn{1}{c}{RMSE}             & \multicolumn{1}{c}{MDE}               & \multicolumn{1}{c}{MAE}             & \multicolumn{1}{c}{RMSE}             & \multicolumn{1}{c}{MDE}               \\ \cline{1-7}
Baseline pre-trained                       & 17.07            & 18.13            & 0.63             & 17.07            & 18.13            & 0.63              \\ 
Baseline fine-tuned                        & 16.41 $\pm$ {\footnotesize 0.24} & 17.50 $\pm$ {\footnotesize 0.23} & 0.60 $\pm$ {\footnotesize 0.010} & 14.33 $\pm$ {\footnotesize 0.16} & 15.55 $\pm$ {\footnotesize 0.15} & 0.54 $\pm$ {\footnotesize 0.006} \\ 
Meta pre-trained             & 16.45 & 16.7 & 0.627 & 16.45 & 16.7 & 0.627 \\
\textbf{Ours w/o ROI}                   & 4.32 $\pm$ {\footnotesize 0.74}  & 5.57 $\pm$ {\footnotesize 0.98}  & 0.15 $\pm$ {\footnotesize 0.022}  & 3.82 $\pm$ {\footnotesize 0.39}  & 4.87 $\pm$ {\footnotesize 0.58}  & 0.14 $\pm$ {\footnotesize 0.012} \\
\textbf{Ours w/ ROI}                   & \textbf{3.08 $\pm$ {\footnotesize 0.13}}  & \textbf{4.16 $\pm$ {\footnotesize 0.23}}  & \textbf{0.12 $\pm$ {\footnotesize 0.005}} & \textbf{3.41 $\pm$ {\footnotesize 0.26}}  & \textbf{4.22 $\pm$ {\footnotesize 0.36}}  & \textbf{0.12 $\pm$ {\footnotesize 0.007}}  \\ \Xhline{2\arrayrulewidth}
\end{tabular}
\caption{Results on the UCSD~\cite{chan2008privacycounting} dataset with $K=1$ and $K=5$ images in the target scene. The meta-training is performed on the WorldExpo'10 training data.}
\label{tab:worldexpo_ucsd}
\end{table*}

\begin{figure*}[!ht]
\centering
\begin{tabular}{ccc}
\includegraphics[height=3.5cm,width=5.0cm]{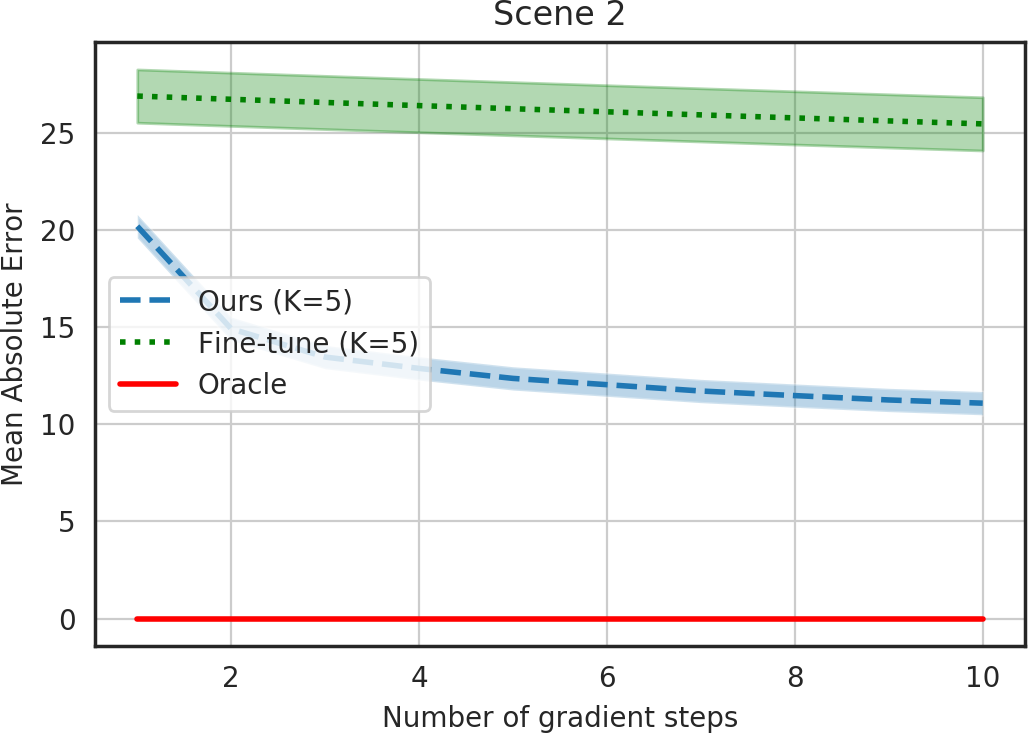} &   \includegraphics[height=3.5cm,width=5.0cm]{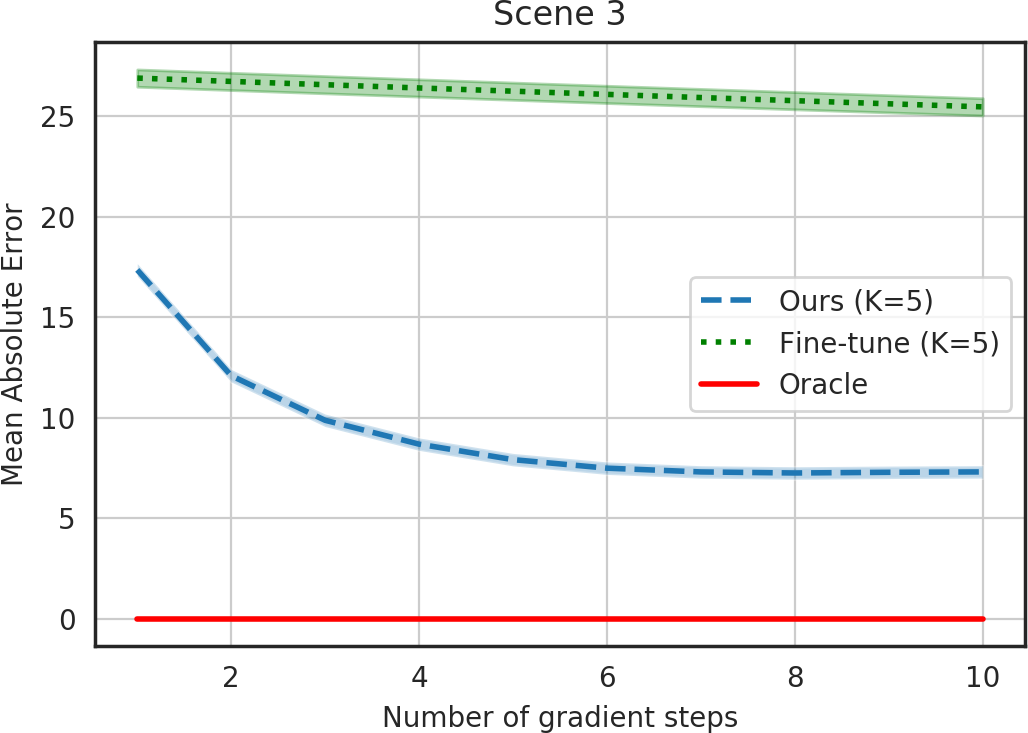} & \includegraphics[height=3.5cm,width=5.0cm]{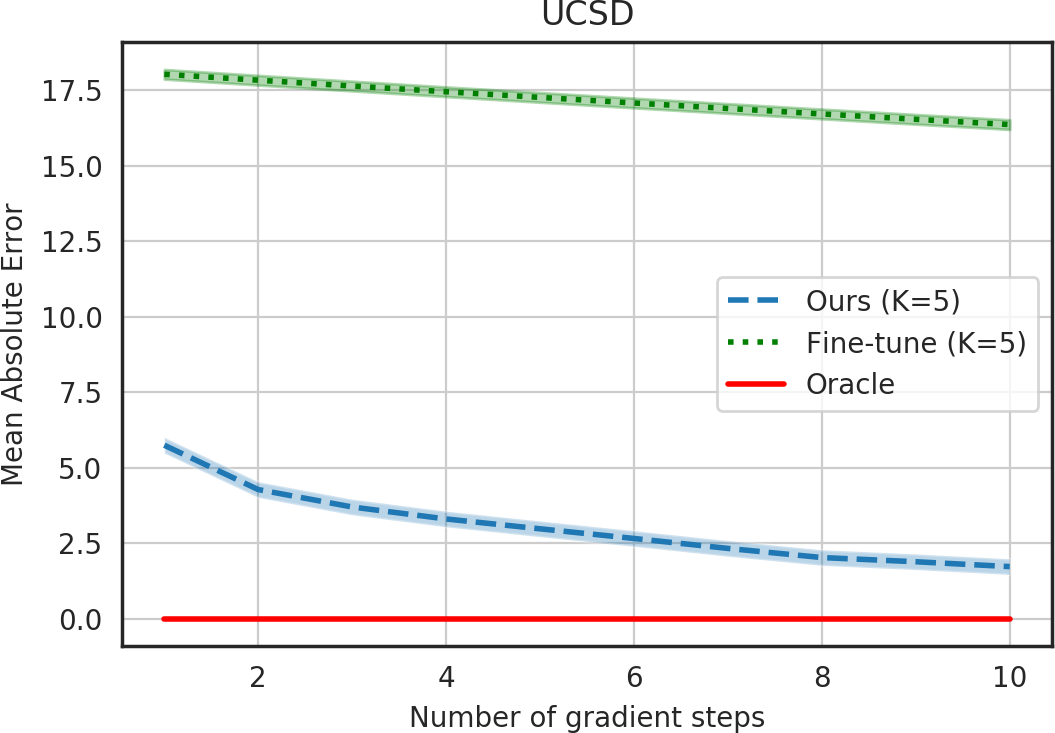} \\
  (a)&(b)&(c) \\ 
\end{tabular}
\caption{Quantitative results of the learning curve during \textit{meta-testing}. The graph (a) shows the learning for Scene 2 and (b) shows the result for Scene 3 in WorldExpo~\cite{zhang2015crosscounting} test sets, respectively. Similarly, (c) shows the learning on UCSD~\cite{chan2008privacycounting}. Note that our approach continues to learn and achieves a lower MAE compared to the baseline fine-tuning approach in ten gradient steps. We consider $K=5$ labeled examples in all three cases. }
\label{fig:fast_adaptation}
\end{figure*}

\noindent \subsection{Datasets and Setup} \label{sec:datasets}

\noindent {\bf Datasets}: Most of the available datasets for crowd-counting are not specifically designed for the scene adaptive crowd counting problem. Our problem formulation requires that the training images are from multiple scenes. To the best of our knowledge, WorldExpo'10~\cite{zhang2015crosscounting} is the only dataset with multiple scenes. We use this dataset for the training of our model. We also consider two other datasets (Mall~\cite{change2013semicounting} and UCSD~\cite{chan2008privacycounting}) for cross-dataset testing. The details of these datasets are described below. \\

The WorldExpo'10~\cite{zhang2015crosscounting} dataset consists of 3980 labeled images from 1132 video sequences based on 108 different scenes. We consider 103 scenes for training and the remaining 5 scenes for testing. The image resolution is fixed at 576 $\times$ 720. When testing on a target scene, we randomly choose $K \in \{1, 5\}$ images from the available images in this scene and use them for obtaining the scene adaptive model parameters $\tilde{\theta}$ (see Fig.~\ref{eq:inner_update}). We then use the remaining images from this scene to calculate the performance of the parameters $\tilde{\theta}$. The Mall~\cite{change2013semicounting} dataset consists of 2000 images from the same camera setup inside a mall. The resolution of each image is 640 $\times$ 480. We follow the standard split, which consists of 800 training images and 1200 test images. Similar to the setup explained earlier, we consider $K \in \{1, 5\}$ images from the training set for fine-tuning the model to obtain the scene adaptive model parameters $\tilde{\theta}$ and later test the model on the test set. The UCSD~\cite{chan2008privacycounting} dataset consists of 2000 images from the same surveillance camera setup to capture a pedestrian scene. The crowd density is relatively sparse, ranging from 11 to 46 persons in an image. The resolution of each image is 238 $\times$ 158. We follow the standard split by considering the first 800 frames for training and 1200 images for testing. We use the same experiment setup of the Mall dataset. 


\noindent {\bf Ground-truth Density Maps}: All datasets come with dot annotations, where each person in the image is annotated with a single point. Following~\cite{li2018csrnetcounting, zhang2016singlecounting}, we use a Gaussian kernel to blur the point annotations in an image to create the ground-truth density map. We set the value of $\sigma$ = 3 in the Gaussian kernel by following \cite{li2018csrnetcounting}. 

\begin{figure*}[!ht]
	\centering
	\begin{tabular}{cc}
		\includegraphics[width=8.5cm,height=2.6cm]{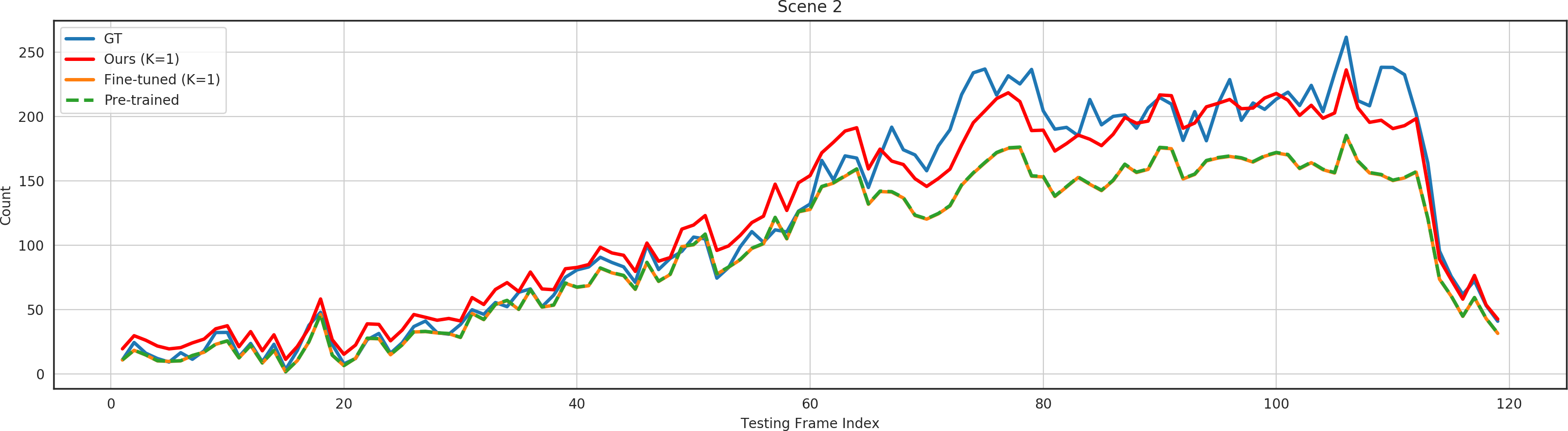} &   \includegraphics[width=8.5cm,height=2.7cm]{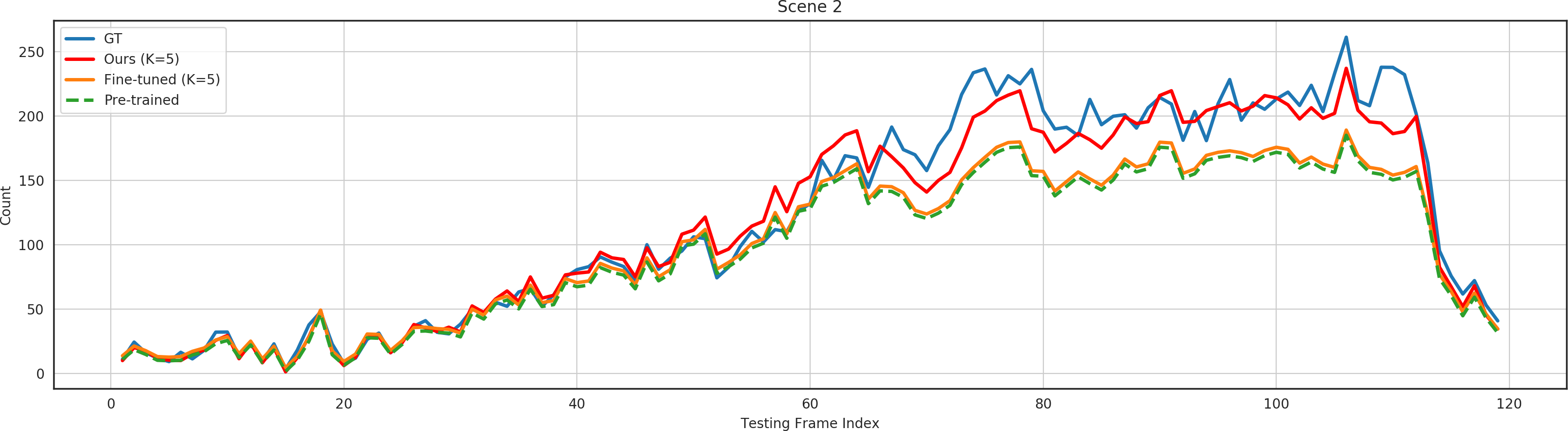} \\
		(a) K=1, Scene 2 & (b) K=5, Scene 2\\ 
		\includegraphics[width=8.5cm,height=2.6cm]
		{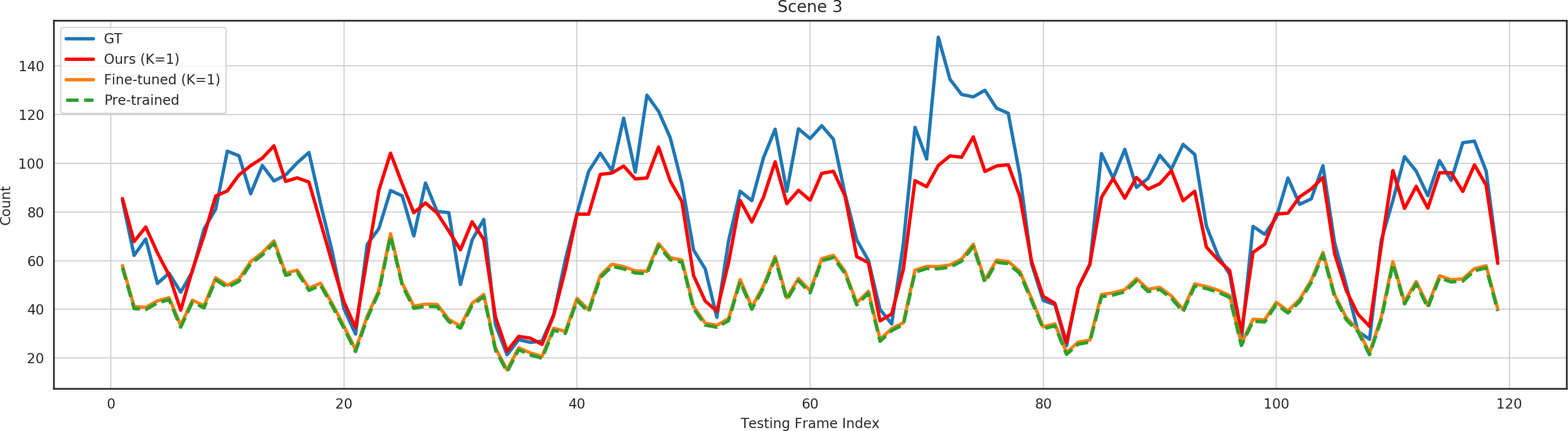} &
		\includegraphics[width=8.5cm,height=2.6cm]{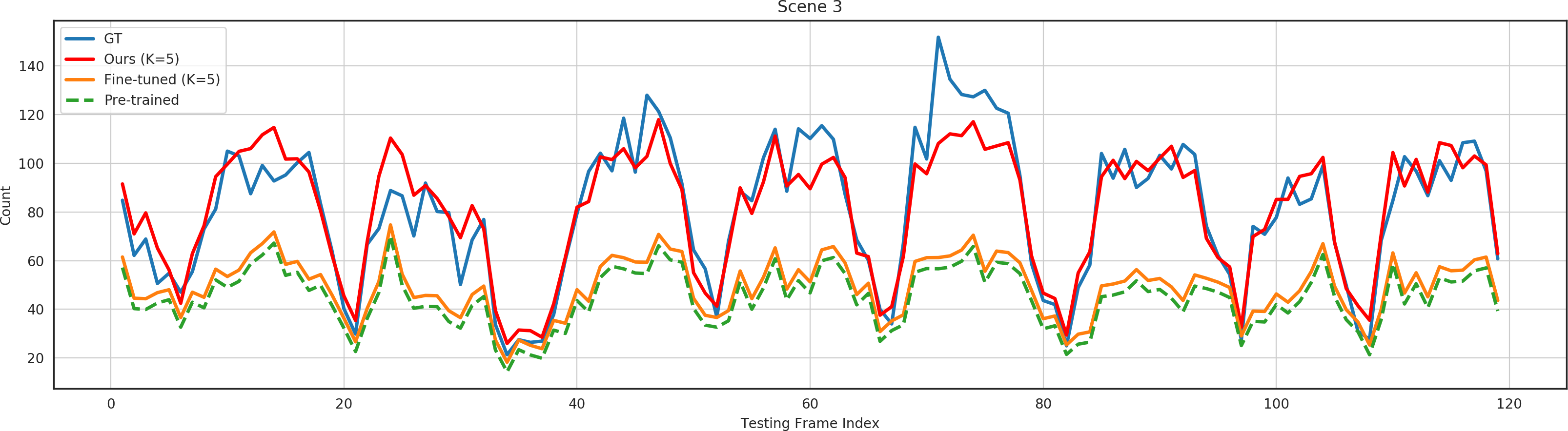} \\
		(c) K=1, Scene 3 & (d) K=5, Scene 3\\
		\includegraphics[width=8.5cm,height=2.6cm]
		{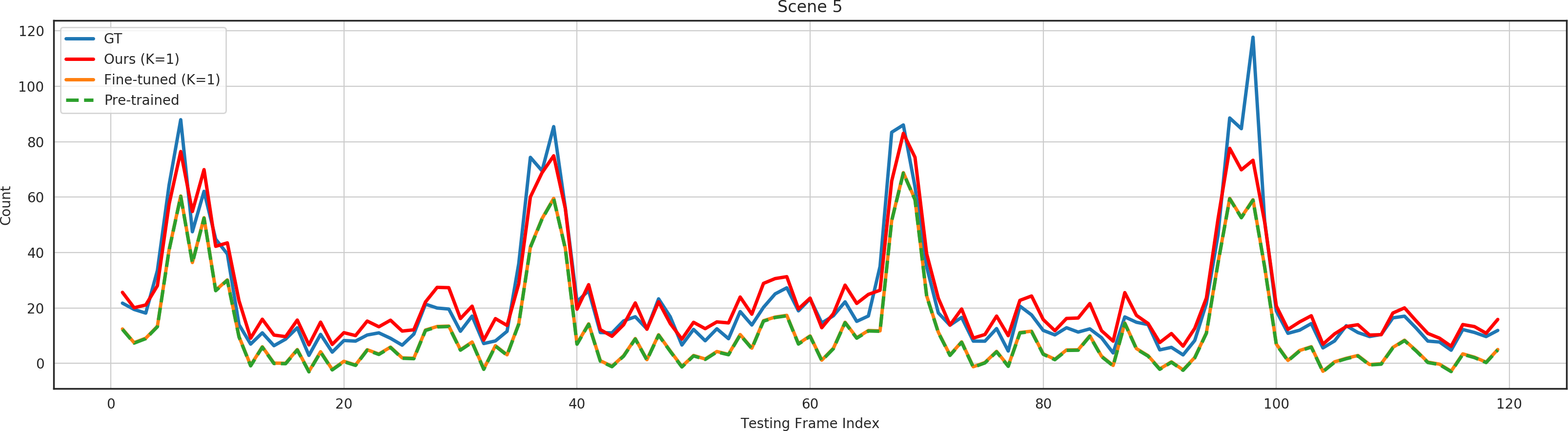} &
		\includegraphics[width=8.5cm,height=2.6cm]{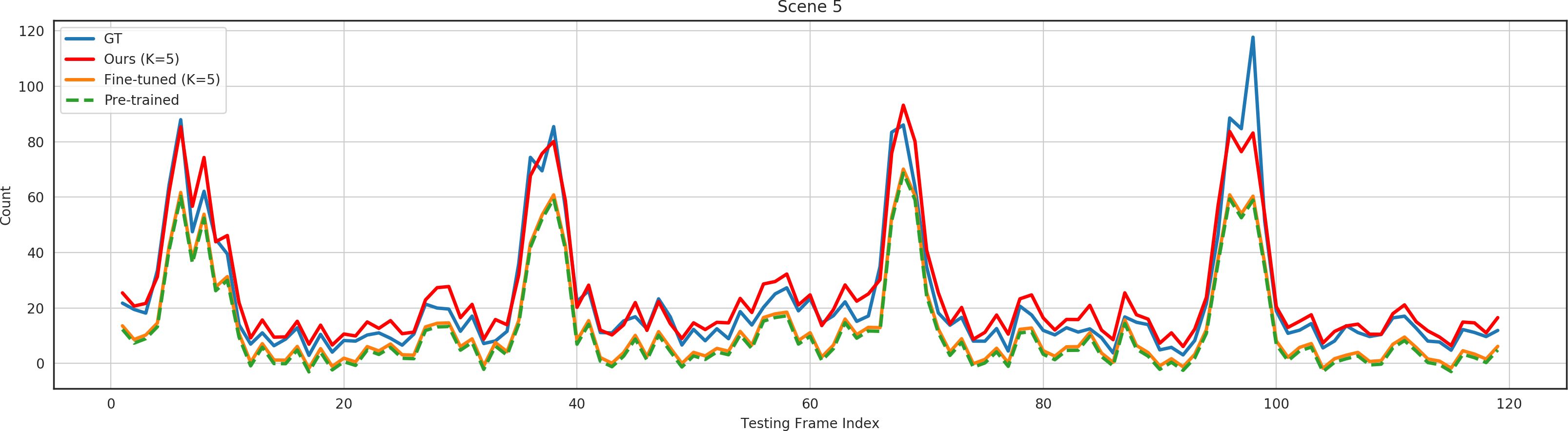} \\
		(e)K=1, Scene 5 & (f) K=5, Scene 5 \\
	\end{tabular}
	\caption{Crowd counting performance comparison between the baselines and our approaches in different scene-specific images from WorldExpo'10~\cite{zhang2015crosscounting} dataset. The labels include, (a) $K=1$ in Scene 2, (b) $K=5$ in Scene 2, (c) $K=1$ in Scene 3, (d) $K=5$ in Scene 3, (e) $K=1$ in Scene 5 and (f) $K=5$ in Scene 5. Note that our approaches outperform the baselines in different settings and is robust to varying crowd density.}
	\label{fig:crowd_count}
\end{figure*}

\noindent {\bf Implementation Details}: We use PyTorch~\cite{paszke2017automatic} for the implementation of our approach. The backbone crowd counting network is implemented based on the source code from the original CSRNet paper~\cite{li2018csrnetcounting}. To generate the \textit{Baseline pre-trained} network, we follow the procedure described in ~\cite{li2018csrnetcounting}. During the meta-learning phase, we initialize the network with baseline pre-trained model. We freeze the feature extractor and only train the density map estimator of the network. We set the hyper-parameters $\alpha = 0.001$ for the inner-update in SGD (see Eq.~\ref{eq:inner_update}) and $\beta = 0.001$ in the outer-update (see Eq.~\ref{eq:outer_update}) in Adam~\cite{kingma2015adamoptim}. We randomly sample a scene for each episode during inner-update. 

\noindent \paragraph{Evaluation Metrics}: To evaluate the results, we use the standard metrics in the context of crowd count estimation. The metrics are: Mean Absolute Error (MAE), Root Mean Squared Error (RMSE) and Mean Deviation Error (MDE) as expressed below:
\begin{eqnarray}
&&MAE = \dfrac{1}{N}\sum_{i=1}^{N}|{\delta}^{\hat{y}}_{i} - {\delta}^{y}_{i}|\\
&&RMSE = \sqrt{\dfrac{1}{N}\sum_{i=1}^{N}|{\delta}^{\hat{y}}_{i} - {\delta}^{y}_{i}|^{2}}\\
&&MDE = \dfrac{1}{N}\sum_{i=1}^{N}\dfrac{|{\delta}^{\hat{y}}_{i} - {\delta}^{y}_{i}|}{{\delta}^{y}_{i}}
\end{eqnarray}
where $N$ is the total number of images in a given camera scene, ${\delta}^{\hat{y}}_{i}$ represents the crowd count of the density map generated by the model and ${\delta}^{y}_{i}$ is the corresponding crowd count of ground-truth density map for the $i$-th input image. Let $p_{h, w}$ be the value at the spatial location $(h, w)$ in a density map for an image $i$, the count $\delta_{i}$ for the image can be expressed $\delta_{i} = \sum_{h=1}^{H}\sum_{w=1}^{W}{p_{h,w}}$, where $H\times W$ is the spatial size of the density map. 

\noindent \subsection{Baselines} \label{sec:baselines}
We define the following baselines for comparison. Note that these baselines have the same backbone architecture as our approach.

\noindent {\bf Baseline pre-trained}: This baseline is a standard crowd counting model as in~\cite{li2018csrnetcounting} trained in a standard supervised setting. The model parameters are trained from all images in the training set. Once the training is done, the model is evaluated directly on images in the new target scene without any adaptation. Note that, the original model in~\cite{li2018csrnetcounting} uses the perspective maps and ground-truth ROI to enhance the final scores, we do not use them for the sake of simplicity.

\noindent {\bf Baseline fine-tuned}: In this baseline, we first consider the \textit{Baseline pre-trained} crowd counting model learned $\theta$ from the standard supervised setting. For a given new scene during testing, we fix the parameters of the feature extractor and fine-tune only the density map estimator using a few images $K \in \{1, 5\}$ from the target scene.

\noindent {\bf Meta pre-trained}: This baseline is similar to our approach, but without the fine-tuning on the target scene. Intuitively, it is similar to ``baseline pre-trained''.


\begin{table*}[!ht]
  \centering
  \footnotesize
  \resizebox{0.9\textwidth}{!}{%
\def\arraystretch{1.1}
\begin{tabular}{cccccccc}
\Xhline{2\arrayrulewidth}
\multirow{2}{*}{\textbf{Target}} & \multirow{2}{*}{\textbf{Methods}} & \multicolumn{3}{c}{\textbf{1-shot (K=1)}}                                         & \multicolumn{3}{c}{\textbf{5-shot (K=5)}}                                         \\ 
                                 &                                   & \multicolumn{1}{c}{MAE}                       & \multicolumn{1}{c}{RMSE}                       & \multicolumn{1}{c}{MDE}                        & \multicolumn{1}{c}{MAE}                       & \multicolumn{1}{c}{RMSE}                       & \multicolumn{1}{c}{MDE}                        \\ \cline{1-8} 
\multirow{4}{*}{WorldExpo (Avg.)} 
& Meta-LSTM~\cite{ravi2016optimizationmethod}             & 13.33 & 18.22 & 0.252 & 12.7 & 16.61 & 0.223 \\
& Reptile~\cite{nichol2018first}             & 11.63 & 15.07 & 0.260 & 8.20 & 11.31 & 0.181 \\
& \textbf{Ours w/o ROI}            & 7.5 & 10.22 & 0.197 & 7.7 & 10.93 & 0.165 \\
& \textbf{Ours w/ ROI}            & \textbf{7.12} & \textbf{9.88} & \textbf{0.172} & \textbf{7.05} & \textbf{9.83} & \textbf{0.155} \\ \hline

\multirow{4}{*}{Mall}
& Meta-LSTM~\cite{ravi2016optimizationmethod}              & 3.95 $\pm$ {\footnotesize 0.04} & 4.34 $\pm$ {\footnotesize 0.537} & 0.12 $\pm$ {\footnotesize 0.002} & 3.54 $\pm$ {\footnotesize 0.44} & 4.41 $\pm$ {\footnotesize 0.472} & 0.10 $\pm$ {\footnotesize 0.014}\\
& Reptile~\cite{nichol2018first}             & 2.55 $\pm$ {\footnotesize 0.07} & 3.26 $\pm$ {\footnotesize 0.09} & 0.079 $\pm$ {\footnotesize 0.001} & 2.49 $\pm$ {\footnotesize 0.23}& 3.20 $\pm$ {\footnotesize 0.29}& 0.078 $\pm$ {\footnotesize 0.006}\\
& \textbf{Ours w/o ROI}            & 2.52 $\pm$ {\footnotesize 0.08} & 3.26 $\pm$ {\footnotesize 0.12} & 0.078 $\pm$ {\footnotesize 0.002}& 2.53 $\pm$ {\footnotesize 0.18}& 3.25 $\pm$ {\footnotesize 0.27}& 0.078 $\pm$ {\footnotesize 0.004}\\
& \textbf{Ours w/ ROI}            & \textbf{2.44 $\pm$ {\footnotesize 0.02}} & \textbf{3.12 $\pm$ {\footnotesize 0.03}} & \textbf{0.076 $\pm$ {\footnotesize 0.001}} & \textbf{2.37 $\pm$ {\footnotesize 0.02}} & \textbf{3.04 $\pm$ {\footnotesize 0.01}} & \textbf{0.073 $\pm$ {\footnotesize 0.001}} \\ \hline

\multirow{4}{*}{UCSD}
& Meta-LSTM~\cite{ravi2016optimizationmethod}             & 14.15 $\pm$ {0.48} &  16.29 $\pm$ {0.425} & 0.463 $\pm$ {0.018} & 13.81 $\pm$ {0.10} & 15.99 $\pm$ {0.009} & 0.45 $\pm$ {0.004} \\
& Reptile~\cite{nichol2018first}             & 5.64 $\pm$ {\footnotesize 2.05} & 6.85 $\pm$ {\footnotesize 2.06} & 0.20 $\pm$ {\footnotesize 0.075} & 4.48 $\pm$ {\footnotesize 0.88}& 5.62 $\pm$ {\footnotesize 0.99} & 0.166 $\pm$ {\footnotesize 0.033}\\
& \textbf{Ours w/o ROI}            & 4.32 $\pm$ {\footnotesize 0.74} & 5.57 $\pm$ {\footnotesize 0.98} & 0.15 $\pm$ {\footnotesize 0.022} & 3.82 $\pm$ {\footnotesize 0.39}& 4.87 $\pm$ {\footnotesize 0.58} & 0.14 $\pm$ {\footnotesize 0.012}\\
& \textbf{Ours w/ ROI}            & \textbf{3.08 $\pm$ {\footnotesize 0.13}} & \textbf{4.16 $\pm$ {\footnotesize 0.23}} & \textbf{0.12 $\pm$ {\footnotesize 0.005}} & \textbf{3.41 $\pm$ {\footnotesize 0.26}} & \textbf{4.22 $\pm$ {\footnotesize 0.36}} & \textbf{0.12 $\pm$ {\footnotesize 0.007}} \\

 \Xhline{2\arrayrulewidth}
\end{tabular}}
\caption{The overall results for adaptation on WorldExpo'10~\cite{zhang2015crosscounting} test set, Mall~\cite{change2013semicounting} and UCSD~\cite{chan2008privacycounting} with $K = 1$ and $K = 5$ train images. We compare with other optimization based meta-learning approaches ``Reptile''~\cite{nichol2018first} and ``Meta-LSTM''~\cite{ravi2016optimizationmethod}.}
\label{tab:meta_overview}
\end{table*}

\begin{table}[!ht]
\centering
\footnotesize
\def\arraystretch{1.1}
\begin{tabular}{cccc}
\Xhline{2\arrayrulewidth}
\multirow{2}{*}{\textbf{Methods}} & \multicolumn{3}{c}{\textbf{1-shot (K=1)}} \\
&	\multicolumn{1}{c}{MAE}	&	\multicolumn{1}{c}{RMSE}	\\	\cline{1-3}
Hossain \textit{et al.}~\cite{hossain2019one}	&	8.23	&	12.08	\\ 
\textbf{Ours w/o ROI}	&	7.5	&	10.22	\\
\textbf{Ours w/ ROI}	&	\textbf{7.12}	&	\textbf{9.88}	\\ \Xhline{2\arrayrulewidth}
\end{tabular}
\caption{Comparison of results on the WorldExpo'10~\cite{zhang2015crosscounting} dataset with $K=1$ images in the target scene with Hossain \textit{et al.}~\cite{hossain2019one}. We use the standing train/test split on WorldExpo'10. Our approach outperforms Hossain \textit{et al.}~\cite{hossain2019one}.}
\label{tab:bmvc}
\end{table}

\noindent \subsection{Experimental Results}\label{sec:experimental_results}

\noindent {\bf Main Results}: Table~\ref{tab:worldexpo} shows the results on the WorldExpo'10 dataset for the 5 test (or target) scenes. We show the results of using both $K=1$ and $K=5$ images for fine-tuning in the test scene. This dataset also comes with ground-truth region-of-interest (ROI). We report the results \textit{with} (\textit{w/}) and \textit{without} (\textit{w/o}) ROI. We repeat the experiments 5 times in each setting with $K$ randomly selected images. We average the scores across the 5 trials and report the standard deviation along with the mean of the scores in Table~\ref{tab:worldexpo}. We report the results from our models as ``\textit{Ours w/o ROI}'' and ``\textit{Ours w/ ROI}''. We compare with the three baselines defined in Sec.~\ref{sec:baselines}. Our models outperform the baselines in most cases. This shows that the meta-learning fine-tuning improves the model's performance. Note that our problem setup requires $K$ labeled images in the test set and hence these $K$ images have to be excluded in the calculation of the evaluation metrics, i.e., we have slightly fewer test images for the results in Table~\ref{tab:worldexpo}. Therefore, the performance numbers in Table~\ref{tab:worldexpo} should not be directly compared with previously reported numbers in the crowd counting literature since our problem formulation is completely different. Besides, some previous crowd counting works~\cite{li2018csrnetcounting} use additional components (e.g., perspective maps) to enhance the final performance. We do not consider these additional components in our models for the sake of simplicity (also the publicly available source code for \cite{li2018csrnetcounting} does not implement those extra components), so the number for ``Baseline pre-trained'' in Table~\ref{tab:worldexpo} is slightly worse than the number reported in \cite{li2018csrnetcounting}.


Table~\ref{tab:worldexpo_mall} and Table~\ref{tab:worldexpo_ucsd} show the results on the Mall and UCSD datasets, respectively. Here we use the training data of WorldExpo'10 for the meta-learning. We then use Mall and UCSD for the scene adaptation and evaluation. This cross-dataset testing can demonstrate the generalization of the proposed method. Our model clearly outperforms the baselines.

 To gain further insights into our method, we visualize the MAE over the number of gradient steps in Fig.~\ref{fig:fast_adaptation} for different scenes. In all the cases, our proposed approach has a better start in learning and improves continuously with more gradient steps. In the three cases shown in Fig.~\ref{fig:fast_adaptation}, our approach performs significantly better that fine-tuning with the same number of gradient updates. 

In Fig.~\ref{fig:crowd_count}, we show the comparison of the crowd count estimations between our approaches and baselines in different scenes in WordExpo'10. Our method consistently produces crowd counts that are closer to the ground-truth compared with other baselines. Some qualitative examples of the density maps generated by our method and baselines are shown in Fig. 1 of the supplementary material.

In Table~\ref{tab:bmvc}, we compare our results with the one-shot scene adaptation proposed in~\cite{hossain2019one} based on the standard WorldExpo'10 data split. In \cite{hossain2019one}, the last two layers in the pre-trained model are fine-tuned to adapt to the target scene. Our approach outperforms \cite{hossain2019one}.


\noindent {\bf Ablation Studies:} Our approach is based on MAML~\cite{finn2017modelagnostic}. In the literature, there are other optimized-based meta-learning approaches, e.g., \cite{nichol2018first,ravi2016optimizationmethod}. We perform additional ablation studies on the effect of different meta-learning frameworks. The results are shown in Table~\ref{tab:meta_overview}. Nichol \textit{et al}.~\cite{nichol2018first} propose an optimization based meta-learning approach similar to~\cite{finn2017modelagnostic} using gradient descent. However, \cite{nichol2018first} differs from \cite{finn2017modelagnostic} in that it does not consider the second-order derivative in the meta-optimization. As a result, its performance is slightly lower as reported in Table~\ref{tab:meta_overview} although it converges faster. In case of \cite{ravi2016optimizationmethod}, the meta-learner is a LSTM based model unlike in~\cite{finn2017modelagnostic} and~\cite{nichol2018first}, because of the similarity between the gradient update in backpropagation and the cell-state update in LSTM. The drawback of~\cite{ravi2016optimizationmethod} is the large number of trainable parameters and different architectures for the learner and meta-learner. In general, the MAML-based meta learning that our approach uses outperforms other optimization-based meta-learning approaches. In Table~\ref{tab:meta_overview}, we highlight only the average score across 5 scenes in WorldExpo due to page limit. For more detailed results of every scene in WorldExpo, refer to Table 1 in the supplementary material. The training scheme for~\cite{nichol2018first, ravi2016optimizationmethod} is similar to our approach based on the same backbone network~\cite{li2018csrnetcounting} as described in the implementation details (Sec~\ref{sec:datasets}). For the hyperparameters setting, please refer to~\cite{nichol2018first, ravi2016optimizationmethod}.

\section{Conclusion}\label{sec:conclusion}

\noindent In this paper, we have addressed the problem of few-shot scene adaptation for crowd counting. We have proposed a meta-learning inspired approach to address the learning mechanism for few-shot scenario. Our proposed approach learns the model parameters in a way that facilitates fast adaptation to new target scenes. Our experimental results show that our proposed approach can learn to quickly adapt to new scenes with only a small number of labeled images from the target camera scene. We believe that our work will help to increase the adoption of crowd counting techniques in the real-world applications.

\noindent{\bf Acknowledgement:} We thank NSERC for funding and NVIDIA for donating some of the GPUs used in this work.

{\small
\bibliographystyle{ieee}
\bibliography{wang}
}
\end{document}